\documentclass{article}
 \usepackage[nonatbib,final]{nips_2017}
\usepackage[utf8]{inputenc} 
\usepackage[T1]{fontenc}    
\usepackage{hyperref}       
\usepackage{url}            
\usepackage{booktabs}       
\usepackage{amsfonts}       
\usepackage{nicefrac}       
\usepackage{microtype}      
\usepackage{times,helvet,courier}
\usepackage{graphicx}
\usepackage[tight,footnotesize]{subfigure}
\usepackage{amsmath,amssymb,amsthm,bm}
\usepackage{multirow,makecell,footmisc}
\usepackage[ruled,vlined]{algorithm2e}
\usepackage{color}

\title{Partial Transfer Learning with Selective Adversarial Networks}

\begin{document}

\author{
Zhangjie Cao$^\dag$, Mingsheng Long$^\dag$, Jianmin Wang$^\dag$, and Michael I. Jordan$^\sharp$\\
$^\dag$KLiss, MOE; TNList; School of Software, Tsinghua University, China\\
$^\sharp$University of California, Berkeley, Berkeley, USA\\
{\tt caozhangjie14@gmail.com\quad\{mingsheng,jimwang\}@tsinghua.edu.cn}\\
{\tt jordan@berkeley.edu}
}

\maketitle

\begin{abstract}
Adversarial learning has been successfully embedded into deep networks to learn transferable features, which reduce distribution discrepancy between the source and target domains. Existing domain adversarial networks assume fully shared label space across domains. In the presence of big data, there is strong motivation of transferring both classification and representation models from existing big domains to unknown small domains. This paper introduces partial transfer learning, which relaxes the shared label space assumption to that the target label space is only a subspace of the source label space. Previous methods typically match the whole source domain to the target domain, which are prone to negative transfer for the partial transfer problem. We present Selective Adversarial Network (SAN), which simultaneously circumvents negative transfer by selecting out the outlier source classes and promotes positive transfer by maximally matching the data distributions in the shared label space. Experiments demonstrate that our models exceed state-of-the-art results for partial transfer learning tasks on several benchmark datasets.
\end{abstract}

\section{Introduction}
Deep networks have significantly improved the state of the art for a wide variety of machine learning problems and applications. At the moment, these impressive gains in performance come only when massive amounts of labeled data are available. Since manual labeling of sufficient training data for diverse application domains on-the-fly is often prohibitive, for problems short of labeled data, there is strong motivation to establishing effective algorithms to reduce the labeling consumption, typically by leveraging off-the-shelf labeled data from a different but related source domain. This promising transfer learning paradigm, however, suffers from the shift in data distributions across different domains, which poses a major obstacle in adapting classification models to target tasks \cite{cite:TKDE10TLSurvey}. 

Existing transfer learning methods assume shared label space and different feature distributions across the source and target domains. These methods bridge different domains by learning domain-invariant feature representations without using target labels, and the classifier learned from source domain can be directly applied to target domain. Recent studies have revealed that deep networks can learn more transferable features for transfer learning \cite{cite:ICML14DeCAF,cite:NIPS14CNN}, by disentangling explanatory factors of variations behind domains. The latest advances have been achieved by embedding transfer learning in the pipeline of deep feature learning to extract domain-invariant deep representations \cite{cite:Arxiv14DDC,cite:ICML15DAN,cite:ICML15RevGrad,cite:ICCV15SDT,cite:NIPS16RTN}.

\begin{figure}[tbp]
  \centering
  \includegraphics[width=0.9\textwidth]{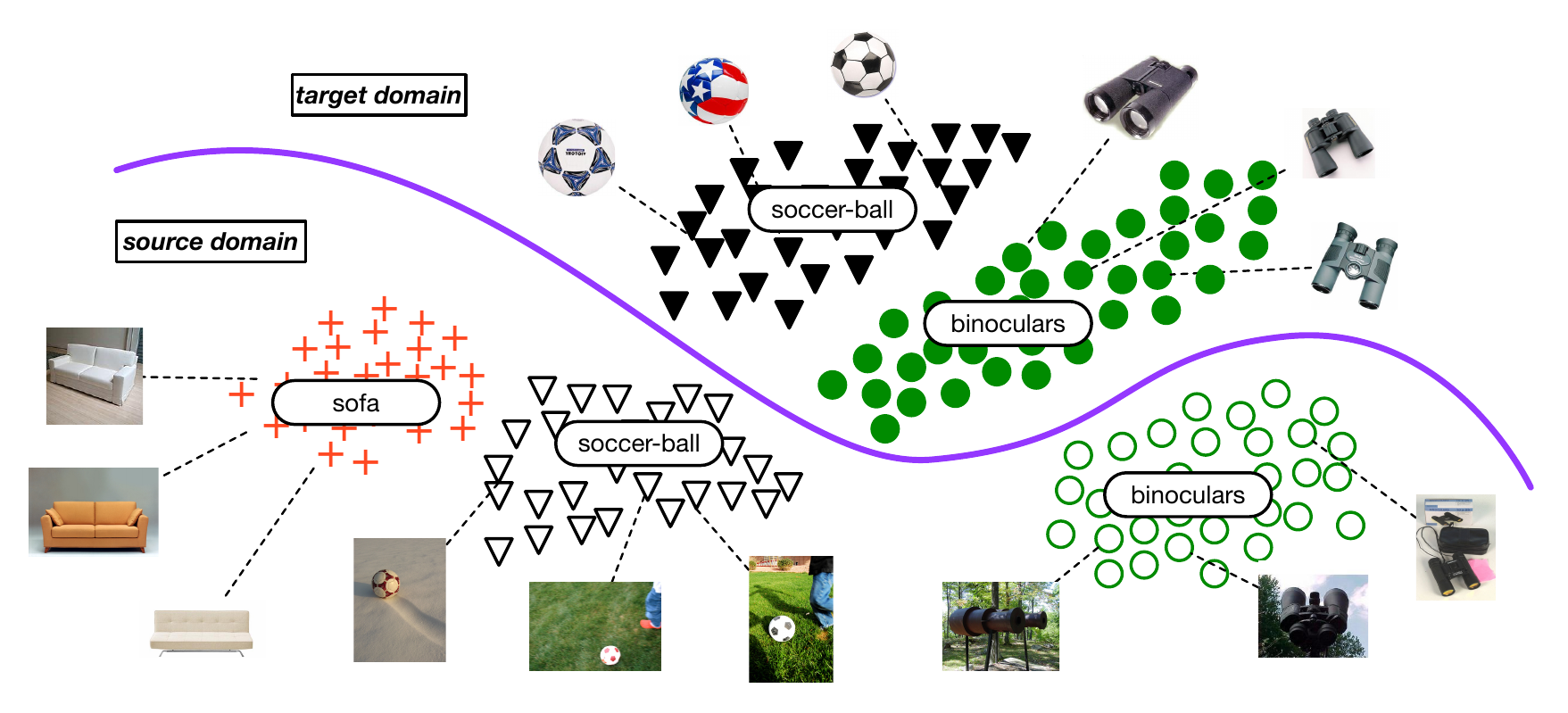}
  \vspace{-10pt}
  \caption{The partial transfer learning problem, where source label space subsumes target label space.}
   \label{fig:SANproblem}
   \vspace{-10pt}
\end{figure}

In the presence of big data, we can readily access large-scale labeled datasets such as ImageNet-1K. Thus, a natural ambition is to directly transfer both the representation and classification models from large-scale dataset to our target dataset, such as Caltech-256, which are usually small-scale and with unknown categories at training and testing time. From big data viewpoint, we can assume that the large-scale dataset is big enough to subsume all categories of the small-scale dataset. Thus, we introduce a novel \emph{partial} transfer learning problem, which assumes that the target label space is a subspace of the source label space. As shown in Figure~\ref{fig:SANproblem}, this new problem is more general and challenging than standard transfer learning, since outlier source classes (``sofa'') will result in negative transfer when discriminating the target classes (``soccer-ball'' and ``binoculars''). Thus, matching the whole source and target domains as previous methods is not an effective solution to this new problem.

This paper presents Selective Adversarial Networks (SAN), which largely extends the ability of deep adversarial adaptation \cite{cite:ICML15RevGrad} to address partial transfer learning from big domains to small domains. SAN aligns the distributions of source and target data in the shared label space and more importantly, selects out the source data in the outlier source classes. A key improvement over previous methods is the capability to simultaneously promote positive transfer of relevant data and alleviate negative transfer of irrelevant data, which can be trained in an end-to-end framework. Experiments show that our models exceed state-of-the-art results for partial transfer learning on public benchmark datasets.

\section{Related Work}
Transfer learning \cite{cite:TKDE10TLSurvey} bridges different domains or tasks to mitigate the burden of manual labeling for machine learning~\cite{cite:TNN11TCA,cite:TPAMI12DTMKL,cite:ICML13TCS,cite:NIPS14FTL}, computer vision \cite{cite:ECCV10Office,cite:CVPR12GFK,cite:NIPS14LSDA} and natural language processing \cite{cite:JMLR11MTLNLP}. The main technical difficulty of transfer learning is to formally reduce the distribution discrepancy across different domains. Deep networks can learn abstract representations that disentangle different explanatory factors of variations behind data \cite{cite:TPAMI13DLSurvey} and manifest invariant factors underlying different populations that transfer well from original tasks to similar novel tasks \cite{cite:NIPS14CNN}. Thus deep networks have been explored for transfer learning \cite{cite:ICML11DADL,cite:CVPR13MidLevel,cite:NIPS14LSDA}, multimodal and multi-task learning \cite{cite:JMLR11MTLNLP,cite:ICML11MDL}, where significant performance gains have been witnessed relative to prior shallow transfer learning methods. 

However, recent advances show that deep networks can learn abstract feature representations that can only reduce, but not remove, the cross-domain discrepancy \cite{cite:ICML11DADL,cite:Arxiv14DDC}, resulting in unbounded risk for target tasks \cite{cite:COLT09DAT,cite:ML10DAT}. Some recent work bridges deep learning and domain adaptation \cite{cite:Arxiv14DDC,cite:ICML15DAN,cite:ICML15RevGrad,cite:ICCV15SDT,cite:NIPS16RTN}, which extends deep convolutional networks (CNNs) to domain adaptation by adding adaptation layers through which the mean embeddings of distributions are matched \cite{cite:Arxiv14DDC,cite:ICML15DAN,cite:NIPS16RTN}, or by adding a subnetwork as domain discriminator while the deep features are learned to confuse the discriminator in a domain-adversarial training paradigm \cite{cite:ICML15RevGrad,cite:ICCV15SDT}. While performance was significantly improved, these state of the art methods may be restricted by the assumption that the source and target domains share the same label space. This assumption is violated in partial transfer learning, which transfers both representation and classification models from existing big domains to unknown small domains. To our knowledge, this is the first work that addresses partial transfer learning in adversarial networks.

\section{Partial Transfer Learning}
In this paper, we propose \emph{partial transfer learning}, a novel transfer learning paradigm where the target domain label space $\mathcal{C}_t$ is a subspace of the source domain label space $\mathcal{C}_s$ i.e. $\mathcal{C}_t \subset \mathcal{C}_s$. This new paradigm finds wide applications in practice, as we usually need to transfer a model from a large-scale dataset (e.g. ImageNet) to a small-scale dataset (e.g. CIFAR10). 
Similar to standard transfer learning, in partial transfer learning we are also provided with a \emph{source} domain $\mathcal{D}_s = \{(\mathbf{x}_i^s,y^s_i)\}_{i=1}^{n_s}$ of $n_s$ labeled examples associated with $|\mathcal{C}_s|$ classes and a \emph{target} domain ${{\mathcal D}_t} = \{ {\mathbf{x}}_i^t\} _{i = 1}^{{n_t}}$ of $n_t$ unlabeled examples associated with $|\mathcal{C}_t|$ classes, but differently, we have $|\mathcal{C}_s| > |\mathcal{C}_t|$ in partial transfer learning. The source domain and target domain are sampled from probability distributions $p$ and $q$ respectively. In standard transfer learning, we have $p \ne q$; and in partial transfer learning, we further have $p_{\mathcal{C}_t} \ne q$, where $p_{\mathcal{C}_t}$ denotes the distribution of the source domain labeled data belonging to label space $\mathcal{C}_t$.
The goal of this paper is to design a deep neural network that enables learning of transfer features $\mathbf{f} = G_f\left( {\bf{x}} \right)$ and adaptive classifier $y = G_y\left( {\bf{f}} \right)$ to bridge the cross-domain discrepancy, such that the target risk ${\Pr _{\left( {{\mathbf{x}},y} \right) \sim q}}\left[ {G_y \left( G_f({\mathbf{x}}) \right) \ne y} \right]$ is minimized by leveraging the source domain supervision.

In standard transfer learning, one of the main challenges is that the target domain has no labeled data and thus the source classifier $G_y$ trained on source domain $\mathcal{D}_s$ cannot be directly applied to target domain $\mathcal{D}_t$ due to the distribution discrepancy of $p \ne q$. In partial transfer learning, another more difficult challenge is that we even do not know which part of the source domain label space $\mathcal{C}_s$ is shared with the target domain label space $\mathcal{C}_t$ because $\mathcal{C}_t$ is not accessible during training, which results in two technical difficulties.
On one hand, the source domain labeled data belonging to \emph{outlier} label space $\mathcal{C}_s \backslash \mathcal{C}_t$ will cause negative transfer effect to the overall transfer performance. 
Existing deep transfer learning methods \cite{cite:ICML15DAN,cite:ICML15RevGrad,cite:ICCV15SDT,cite:NIPS16RTN} generally assume source domain and target domain have the same label space and match the whole distributions $p$ and $q$, which are prone to negative transfer since the source and target label spaces are different and thus cannot be matched in principle.
Thus, how to eliminate or at least decrease the influence of the source labeled data in outlier label space $\mathcal{C}_s \backslash \mathcal{C}_t$ is the key to alleviating negative transfer.
On the other hand, reducing the distribution discrepancy between $p_{\mathcal{C}_t}$ and $q$ is crucial to enabling knowledge transfer in the shared label space $\mathcal{C}_t$.

In summary, there are two essential challenges to enabling partial transfer learning. \textbf{(1)} Circumvent negative transfer by filtering out the unrelated source labeled data belonging to the outlier label space $\mathcal{C}_s \backslash \mathcal{C}_t$. \textbf{(2)} Promote positive transfer by maximally matching the data distributions $p_{\mathcal{C}_t}$ and $q$ in the shared label space $\mathcal{C}_t$. We propose a novel selective adversarial network to address both challenges.

\subsection{Domain Adversarial Network}
Domain adversarial networks have been successfully applied to transfer learning~\cite{cite:ICML15RevGrad,cite:ICCV15SDT} by extracting transferable features that can reduce the distribution shift between the source domain and the target domain. The adversarial learning procedure is a two-player game, where the first player is the domain discriminator $G_d$ trained to distinguish the source domain from the target domain, and the second player is the feature extractor $G_f$ fine-tuned simultaneously to confuse the domain discriminator.


To extract domain-invariant features $\mathbf{f}$, the parameters $\theta_f$ of feature extractor $G_f$ are learned by maximizing the loss of domain discriminator $G_d$, while the parameters $\theta_d$ of domain discriminator $G_d$ are learned by minimizing the loss of the domain discriminator. In addition, the loss of label predictor $G_y$ is also minimized. The objective of domain adversarial network \cite{cite:ICML15RevGrad} is the functional:
\begin{equation}\label{eqn:GRL}
	C_{0} \left( {{\theta _f},{\theta _y},{\theta _d}} \right) = \frac{1}{{{n_s}}}\sum\limits_{{{\mathbf{x}}_i} \in {\mathcal{D}_s}} {{L_y}\left( {{G_y}\left( {{G_f}\left( {{{\mathbf{x}}_i}} \right)} \right),{y_i}} \right)}  - \frac{\lambda }{{{n_s} + {n_t}}}\sum\limits_{{{\mathbf{x}}_i} \in \left( {{\mathcal{D}_s} \cup {\mathcal{D}_t}} \right)} {{L_d}\left( {{G_d}\left( {{G_f}\left( {{{\mathbf{x}}_i}} \right)} \right),{d_i}} \right)} ,
\end{equation}
where $\lambda$ is a trade-off parameter between the two objectives that shape the features during learning.
After training convergence, the parameters $\hat\theta_f$, $\hat\theta_y$, $\hat\theta_d$ will deliver a saddle point of the functional~\eqref{eqn:GRL}: 
\begin{equation}\label{eqn:param1}
\begin{gathered}
     (\hat\theta_f, \hat\theta_y) = \arg \mathop {\min }\limits_{{\theta _f},{\theta _y}} C_0 \left( {{\theta _f},{\theta _y},{\theta _d}} \right), \\
     (\hat\theta_d) = \arg \mathop {\max }\limits_{{\theta_d}} C_0 \left( {{\theta _f},{\theta _y},{\theta _d}} \right).
\end{gathered}
\end{equation}
Domain adversarial networks are among the top-performing architectures for standard transfer learning where the source domain label space and target domain label space are the same, $\mathcal{C}_s = \mathcal{C}_t$.

\begin{figure}[tbp]
  \centering
  \includegraphics[width=0.9\textwidth]{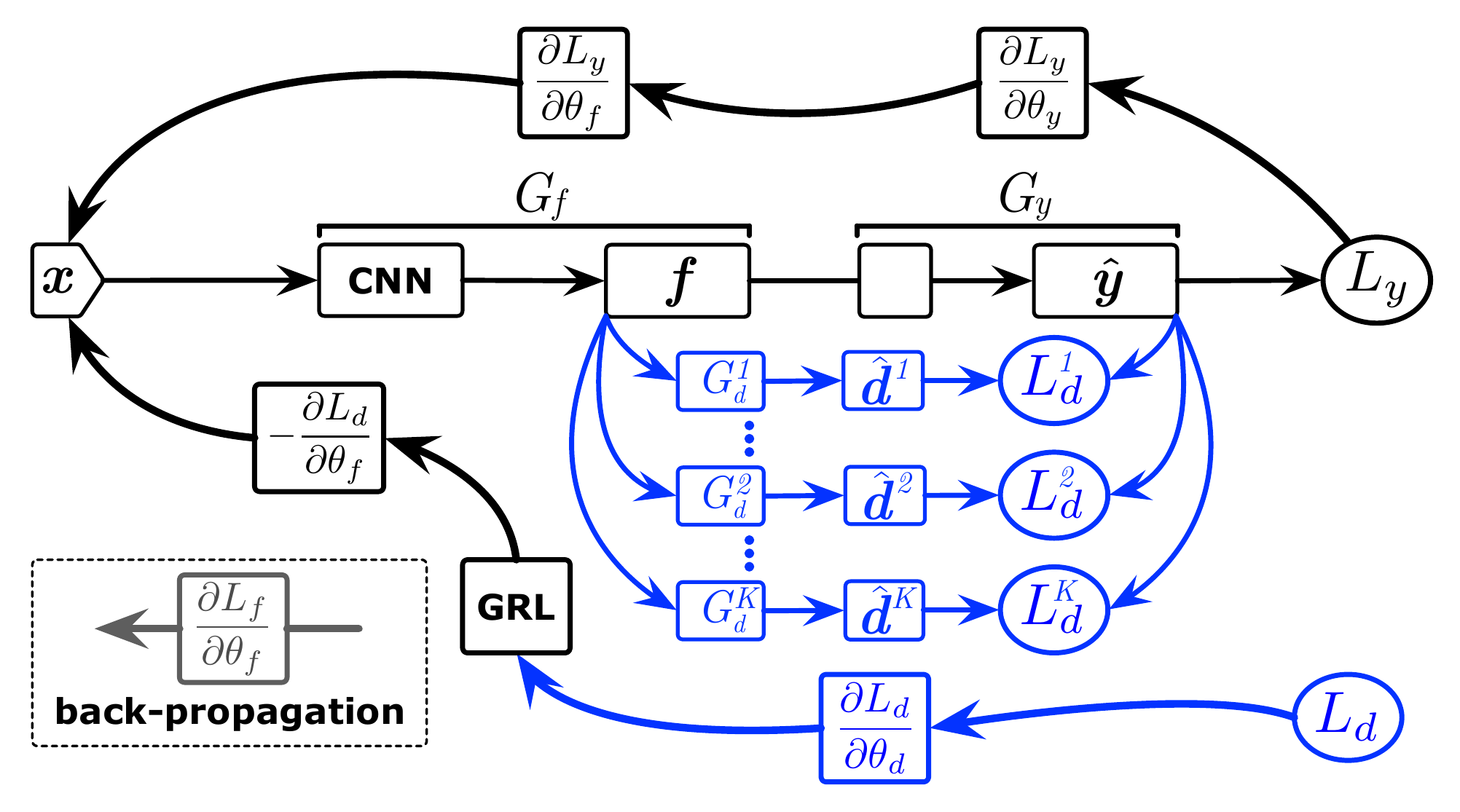}
  \vspace{-10pt}
  \caption{The architecture of the proposed Selective Adversarial Networks (SAN) for partial transfer learning, where $\mathbf{f}$ is the extracted deep features, ${\hat{\mathbf{y}}}$ is the predicted data label, and ${\hat{\mathbf{d}}}$ is the predicted domain label; $G_f$ is the feature extractor, $G_y$ and $L_y$ are the label predictor and its loss, $G_d^k$ and $L_d^k$ are the domain discriminator and its loss; GRL stands for Gradient Reversal Layer. The blue part shows the class-wise adversarial networks ($|\mathcal{C}_s|$ in total) designed in this paper. \emph{Best viewed in color.}}
   \label{fig:SAN}
   \vspace{-10pt}
\end{figure}

\subsection{Selective Adversarial Network}
In partial transfer learning, however, the target domain label space is a subset of the source domain label space, $\mathcal{C}_t \subset \mathcal{C}_s$. Thus, matching the whole source domain distribution $p$ and target domain distribution $q$ will result in negative transfer caused by the \emph{outlier} label space $\mathcal{C}_s \backslash \mathcal{C}_t$. The larger the outlier label space $\mathcal{C}_s \backslash \mathcal{C}_t$ compared to the target label space $\mathcal{C}_t$, the severer the negative transfer effect will be. To combat \emph{negative transfer}, we should find a way to select out the outlier source classes as well as the associated source labeled data in $\mathcal{C}_s \backslash \mathcal{C}_t$ when performing domain adversarial adaptation.

To match the source and target domains of different label spaces $\mathcal{C}_s \ne \mathcal{C}_t$, we need to split the domain discriminator $G_d$ in Equation~\eqref{eqn:GRL} into $|\mathcal{C}_s|$ class-wise domain discriminators $G_d^k, {k=1, \ldots, |\mathcal{C}_s|}$, each is responsible for matching the source and target domain data associated with label $k$, as shown in Figure~\ref{fig:SAN}. Since the target label space $\mathcal{C}_t$ is inaccessible during training while the target domain data are fully unlabeled, it is not easy to decide which domain discriminator $G_d^k$ is responsible for each target data point. Fortunately, we observe that the output of the label predictor ${\hat{\bf{y}}}_i = G_y(\mathbf{x}_i)$ to each data point $\mathbf{x}_i$ is a probability distribution over the source label space $\mathcal{C}_s$. This distribution well characterizes the probability of assigning $\mathbf{x}_i$ to each of the $|\mathcal{C}_s|$ classes. Therefore, it is natural to use ${\hat{\bf{y}}}_i$ as the probability to assign each data point $\mathbf{x}_i$ to the $|\mathcal{C}_s|$ domain discriminators $G_d^k, {k=1, \ldots, |\mathcal{C}_s|}$. The assignment of each point $\mathbf{x}_i$ to different discriminators can be implemented by a \emph{probability-weighted} domain discriminator loss for all $|\mathcal{C}_s|$ domain discriminators $G_d^k, {k=1, \ldots, |\mathcal{C}_s|}$ as follows,
\begin{equation}\label{eqn:Ld}
	{{L'_d}} = \frac{1}{{{n_s} + {n_t}}}\sum\limits_{k = 1}^{|{\mathcal{C}_s}|} {\sum\limits_{{\mathbf{x}_i} \in  {{\mathcal{D}_s} \cup {\mathcal{D}_t}}} {\hat y_i^kL_d^k\left( {G_d^k\left( {{G_f}\left( {{{\mathbf{x}}_i}} \right)} \right),d_i} \right)} } ,
\end{equation}
where $G_d^k$ is the $k$-th domain discriminator while $L_d^k$ is its cross-entropy loss, and $d_i$ is the domain label of point $\mathbf{x}_i$. Compared with the single-discriminator domain adversarial network in Equation~\eqref{eqn:GRL}, the proposed multi-discriminator domain adversarial network enables fine-grained adaptation where each data point $\mathbf{x}_i$ is matched only by those relevant domain discriminators according to its probability ${\hat{\bf{y}}}_i$. This fine-grained adaptation may introduce three benefits. \textbf{(1)} It avoids the hard assignment of each point to only one domain discriminator, which tends to be inaccurate for target domain data. \textbf{(2)} It circumvents negative transfer since each point is only aligned to one or several most relevant classes, while the irrelevant classes are filtered out by the probability-weighted domain discriminator loss. \textbf{(3)} The probability-weighted domain discriminator loss puts different losses to different domain discriminators, which naturally learns multiple domain discriminators with different parameters $\theta_d^k$; these domain discriminators with different parameters can promote \emph{positive transfer} for each instance.

Besides the \emph{instance-level} weighting mechanism described above, we introduce another \emph{class-level} weighting method to further remove the negative influence of outlier source classes $\mathcal{C}_s \backslash \mathcal{C}_t$ and the associated source data. We observe that only the domain discriminators responsible for the target classes $\mathcal{C}_t$ are effective for promoting \emph{positive transfer}, while the other discriminators responsible for the outlier source classes $\mathcal{C}_s \backslash \mathcal{C}_t$ only introduce noises and deteriorate the positive transfer between the source domain and the target domain in the shared label space $\mathcal{C}_t$. Therefore, we need to down-weight the domain discriminators responsible for the outlier source classes, which can be implemented by class-level weighting of these domain discriminators.
Since target data are not likely to belong to the outlier source classes, their probabilities $y_i^k, k \in \mathcal{C}_s \backslash \mathcal{C}_t$ are also sufficiently small. Thus, we can down-weight the domain discriminators responsible for the outlier source classes as follows,
\begin{equation}\label{eqn:Ld2}
	{L_d} = \frac{1}{{{n_s} + {n_t}}}\sum\limits_{k = 1}^{|{\mathcal{C}_s}|} {\left( {\frac{1}{{{n_t}}}\sum\limits_{{{\mathbf{x}}_i} \in {\mathcal{D}_t}} {\hat y_i^k} } \right)\sum\limits_{{\mathbf{x}_i} \in \left( {{\mathcal{D}_s} \cup {\mathcal{D}_t}} \right)} {\hat y_i^kL_d^k\left( {G_d^k\left( {{G_f}\left( {{{\mathbf{x}}_i}} \right)} \right),d_i} \right)} } ,
\end{equation}
where ${\frac{1}{{{n_t}}}\sum\nolimits_{{{\mathbf{x}}_i} \in {\mathcal{D}_t}} {\hat y_i^k} }$ is the class-level weight for class $k$, which is small for the outlier source classes.

Although the multiple domain discriminators introduced in Equation~\eqref{eqn:Ld2} can selectively transfer relevant knowledge to target domain by decreasing the negative influence of outlier source classes $\mathcal{C}_s \backslash \mathcal{C}_t$ and by effectively transferring knowledge of shared label space $\mathcal{C}_t$, it highly depends on the probability ${{\hat{\mathbf{y}}}_i} = G_y(\mathbf{x}_i)$.
Thus, we further refine the label predictor $G_y$ by exploiting the entropy minimization principle~\cite{cite:NIPS04SSLEM} which encourages low-density separation between classes. This criterion is implemented  by minimizing the conditional-entropy $E$ of probability $\hat y_i^k$ on target domain $\mathcal{D}_t$ as
\begin{equation}\label{eqn:Entropy}
	E = \frac{1}{{{n_t}}}\sum\limits_{{{\mathbf{x}}_i} \in {\mathcal{D}_t}} {H\left( {{G_y}\left( {{G_f}\left( {{{\mathbf{x}}_i}} \right)} \right)} \right)} 
\end{equation}
where $H(\cdot)$ is the conditional-entropy loss functional $ H\left( {{G_y}\left( {{G_f}\left( {{{\mathbf{x}}_i}} \right)} \right)} \right) =  - \sum\nolimits_{k = 1}^{|{\mathcal{C}_s}|} {\hat y_i^k\log \hat y_i^k} $. By minimizing entropy~\eqref{eqn:Entropy}, the label predictor $G_y(\mathbf{x}_i)$ can directly access target unlabeled data and will amend itself to pass through the target low-density regions to give more accurate probability ${{\hat{\mathbf{y}}}_i}$.

Integrating all things together, the objective of the proposed Selective Adversarial Network (SAN) is
\begin{equation}\label{eqn:MultiA}
\begin{aligned}
  C\left( {{\theta _f},{\theta _y},\theta _d^k|_{k = 1}^{|{\mathcal{C}_s}|}} \right) &= \frac{1}{{{n_s}}}\sum\limits_{{\mathbf{x}_i} \in {\mathcal{D}_s}} {{L_y}\left( {{G_y}\left( {{G_f}\left( {{\mathbf{x}_i}} \right)} \right)}, y_i \right)}  + \frac{1}{{{n_t}}}\sum\limits_{{{\mathbf{x}}_i} \in {\mathcal{D}_t}} {H\left( {{G_y}\left( {{G_f}\left( {{{\mathbf{x}}_i}} \right)} \right)} \right)}  \\
   &- \frac{\lambda}{{{n_s} + {n_t}}}\sum\limits_{k = 1}^{|{\mathcal{C}_s}|} {\left( {\frac{1}{{{n_t}}}\sum\limits_{{{\mathbf{x}}_i} \in {\mathcal{D}_t}} {\hat y_i^k} } \right)\sum\limits_{{\mathbf{x}_i} \in {{\mathcal{D}_s} \cup {\mathcal{D}_t}} } {\hat y_i^kL_d^k\left( {G_d^k\left( {{G_f}\left( {{{\mathbf{x}}_i}} \right)} \right),d_i} \right)} },  \\ 
\end{aligned}
\end{equation}  
where $\lambda$ is a hyper-parameter that trade-offs the two objectives in the unified optimization problem. The optimization problem is to find the parameters ${\hat\theta_f}$, ${\hat\theta_y}$ and ${\hat\theta_d^k}(k=1,2,...,|\mathcal{C}_s|)$ that satisfy
\begin{equation}\label{eqn:parameter1}
\begin{gathered}
     ({\hat\theta_f}, {\hat\theta_y}) = \arg \mathop {\min }\limits_{{\theta _f},{\theta _y}} C\left( {{\theta _f},{\theta _y},\theta _d^k|_{k = 1}^{|{\mathcal{C}_s}|}} \right), \\
     ({\hat\theta_d^1},...,{\hat\theta_d^{|\mathcal{C}_s|}}) = \arg \mathop {\max }\limits_{{\theta_d^1},...,{\theta_d^{|{\mathcal{C}_s}|}}} C\left( {{\theta _f},{\theta _y},\theta _d^k|_{k = 1}^{|{\mathcal{C}_s}|}} \right).
\end{gathered}
\end{equation}
The selective adversarial network (SAN) successfully enables partial transfer learning, which simultaneously circumvents negative transfer by filtering out outlier source classes $\mathcal{C}_s \backslash \mathcal{C}_t$, and promotes positive transfer by maximally matching the data distributions $p_{\mathcal{C}_t}$ and $q$ in the shared label space $\mathcal{C}_t$.

\section{Experiments}
We conduct experiments on three benchmark datasets to evaluate the efficacy of our approach against several state-of-the-art deep transfer learning methods. Codes and datasets will be available online.

\subsection{Setup}
The evaluation is conducted on three public datasets: Office-31, Caltech-Office and ImageNet-Caltech.

\textbf{Office-31}~\cite{cite:ECCV10Office} is a standard benchmark for domain adaptation in computer vision, consisting of 4,652
images and 31 categories collected from three distinct domains: \textit{Amazon} (\textbf{A}), which contains images
downloaded from amazon.com, \textit{Webcam} (\textbf{W}) and \textit{DSLR} (\textbf{D}), which contain images taken by web
camera and digital SLR camera with different settings, respectively. We denote the three domains with 31 categories as \textbf{A\ 31}, \textbf{W\ 31} and \textbf{D\ 31}. Then we use the ten categories shared by \textit{Office-31} and \textit{Caltech-256} and select images of these ten categories in each domain of \textit{Office-31} as target domains, denoted as \textbf{A\ 10}, \textbf{W\ 10} and \textbf{D\ 10}. We evaluate all methods across
six transfer tasks \textbf{A\ 31} $\rightarrow$ \textbf{W\ 10}, \textbf{D\ 31} $\rightarrow$ \textbf{W\ 10}, \textbf{W\ 31} $\rightarrow$ \textbf{D\ 10}, \textbf{A\ 31} $\rightarrow$ \textbf{D\ 10}, \textbf{D\ 31} $\rightarrow$ \textbf{A\ 10} and \textbf{W\ 31} $\rightarrow$ \textbf{A\ 10}. These tasks represent the performance on the setting where both source and target domains have small number of classes.

\textbf{Caltech-Office}~\cite{cite:CVPR12GFK} is built by using \textit{Caltech-256} (\textbf{C\ 256})~\cite{cite:TR07Caltech} as source domain and the three domains in \textit{Office 31} as target domains. We use the ten categories shared by \textit{Caltech-256} and \textit{Office-31} and select images of these ten categories in each domain of \textit{Office-31} as target domains \cite{cite:CVPR12GFK,cite:ICCV13JDA,cite:AAAI16CORAL}. Denoting source domains as \textbf{C 256}, we can build 3 transfer tasks: \textbf{C 256} $\rightarrow$ \textbf{W 10}, \textbf{C 256} $\rightarrow$ \textbf{A 10} and \textbf{C 256} $\rightarrow$ \textbf{D 10}. This setting aims to test the performance of different methods on the task setting where source domain has much more classes than the target domain.

\textbf{ImageNet-Caltech} is constructed with \textit{ImageNet-1K}~\cite{cite:ILSVRC15} dataset containing 1000 classes and \textit{Caltech-256} containing 256 classes. They share 84 common classes, thus we form two transfer learning tasks: \textbf{ImageNet\ 1000} $\rightarrow$ \textbf{Caltech\ 84} and \textbf{Caltech\ 256} $\rightarrow$ \textbf{ImageNet\ 84}. To prevent the effect of the pre-trained model on ImageNet, we use ImageNet validation set when ImageNet is used as target domain and ImageNet training set when ImageNet is used as source domain. This setting represents the performance on tasks with large number of classes in both source and target domains.

We compare the performance of \textbf{SAN} with state of the art transfer learning and deep learning methods: Convolutional Neural Network (\textbf{AlexNet}~\cite{cite:NIPS12CNN}), Deep Adaptation
Network (\textbf{DAN})~\cite{cite:ICML15DAN}, Reverse Gradient (\textbf{RevGrad})~\cite{cite:ICML15RevGrad} and Residual
Transfer Networks (\textbf{RTN})~\cite{cite:NIPS16RTN}. DAN learns transferable features by embedding deep features of multiple task-specific layers to reproducing kernel Hilbert spaces (RKHSs) and matching different distributions optimally using multi-kernel MMD. RevGrad improves domain adaptation by making the source and target domains indistinguishable for a discriminative domain classifier via an adversarial training paradigm. RTN jointly learns transferable features and adapts different source and target classifiers via deep residual learning \cite{cite:CVPR16DRL}. All prior methods do not address partial transfer learning where the target label space is a subspace of the source label space.
To go deeper with the efficacy of selective mechanism and entropy minimization, we perform ablation study by evaluating two variants of \textbf{SAN}: (1) \textbf{SAN-selective} is the variant without selective mechanism; (2) \textbf{SAN-entropy} is the variant without entropy minimization.

We follow standard protocols and use all labeled source data and all unlabeled target data for unsupervised transfer learning~\cite{cite:ECCV10Office,cite:ICML15DAN}. We compare average classification accuracy of each transfer task using three random experiments. For MMD-based methods (DAN and RTN), we use Gaussian kernel with bandwidth b set to median pairwise squared distances on training data, i.e. median heuristic~\cite{cite:NIPS12MKMMD}. 
For all methods, we perform cross-valuation on labeled source data to select parameters.

We implement all deep methods based on the Caffe deep-learning framework, and fine-tune from Caffe-provided models of AlexNet~\cite{cite:NIPS12CNN} pre-trained on ImageNet. We add a bottleneck layer between the $fc7$ and $fc8$ layers as RevGrad \cite{cite:ICML15RevGrad} except for the task \textbf{ImageNet\ 1000} $\rightarrow$ \textbf{Caltech\ 84} since the pre-trained model is trained on ImageNet dataset and it can fully exploit the advantage of pre-trained model with the original $fc7$ and $fc8$ layer. For SAN, we fine-tune all the feature layers and train the bottleneck layer, the classifier layer and the adversarial networks. Since these new layers and networks are trained from scratch, we set their learning rate to be 10 times that of the other layers. We use mini-batch stochastic gradient descent (SGD) with momentum of 0.9 and the learning rate annealing strategy implemented in RevGrad~\cite{cite:ICML15RevGrad}: the learning rate is not selected through a grid search due to high computational cost: it is adjusted during SGD using the following formula: $\eta_p = \frac{\eta_0}{{(1+\alpha p)}^\beta}$, where $p$ is the training progress linearly changing from 0 to 1, $\eta_0 = 0.001$, $\alpha = 10$ and $\beta = 0.75$, which is optimized for low error on the source domain. As \textbf{SAN} can work stably across different transfer tasks, the penalty of adversarial networks is increased from $0$ to $1$ gradually as RevGrad \cite{cite:ICML15RevGrad}. 

\vspace{-10pt}
\subsection{Results}

The classification results on the six tasks of \textit{Office-31}, the three tasks of \textit{Caltech-Office} and the two tasks of \textit{ImageNet-Caltech} are shown in Table~\ref{table:accuracy_office} and ~\ref{table:accuracy_coic}. The SAN model outperforms all comparison methods on all the tasks. In particular, SAN substantially improves the accuracy by huge margins on tasks with small source domain and small target domain, e.g. \textbf{A 31} $\rightarrow$ \textbf{W 10} , \textbf{A 31} $\rightarrow$ \textbf{D 10}, and  tasks with large source domain and small target domain, e.g. \textbf{C 31} $\rightarrow$ \textbf{W 10}. And it achieves considerable accuracy gains on tasks with large-scale source domain and target domain, e.g. \textbf{I 1000} $\rightarrow$ \textbf{C 84}. These results suggest that SAN can learn transferable features for partial transfer learning in all the tasks under the setting where the target label space is a subspace of the source label space.

\begin{table}[tbp]
    \addtolength{\tabcolsep}{0pt} 
    \centering 
    \caption{Accuracy (\%) of partial transfer learning tasks on \emph{Office-31}}
    \label{table:accuracy_office}
    \begin{scriptsize}
    \vspace{0pt}
    \begin{tabular}{|c|cccccc|c|}
    	\hline
        \multirow{2}{30pt}{\centering Method} & \multicolumn{7}{c|}{Office-31} \\
        \cline{2-8}
        & A 31 $\rightarrow$ W 10 & D 31 $\rightarrow$ W 10 & W 31 $\rightarrow$ D 10 & A 31 $\rightarrow$ D 10 & D 31 $\rightarrow$ A 10 & W 31 $\rightarrow$ A 10 & Avg \\
        \hline
        AlexNet~\cite{cite:NIPS12CNN} & 58.51 & 95.05 & 98.08 & 71.23 & 70.6 & 67.74 & 76.87 \\
        DAN~\cite{cite:ICML15DAN} & 56.52 & 71.86 & 86.78 & 51.86 & 50.42 & 52.29 & 61.62 \\
        RevGrad~\cite{cite:ICML15RevGrad} & 49.49 & 93.55 & 90.44 & 49.68 & 46.72 & 48.81 & 63.11 \\
        RTN~\cite{cite:NIPS16RTN} & 66.78 & 86.77 & 99.36 & 70.06 & 73.52 & 76.41 & 78.82 \\
        \hline
        SAN-selective & 71.51 & 98.31 & 100.00 & 78.34 & 77.87 & 76.32 & 83.73 \\
        SAN-entropy & 74.61 & 98.31 & 100.00 & 80.29 & 78.39 & 82.25 & 85.64 \\ 
        SAN & \textbf{80.02} & \textbf{98.64} & \textbf{100.00} & \textbf{81.28} & \textbf{80.58} & \textbf{83.09} & \textbf{87.27} \\ 
    	\hline
    \end{tabular}
    \end{scriptsize}
    \vspace{-10pt}
\end{table}

\begin{table}[tbp]
    \addtolength{\tabcolsep}{0pt} 
    \centering 
    \caption{Accuracy (\%) of partial transfer learning tasks on \emph{Caltech-Office} and \emph{ImageNet-Caltech}}
    \label{table:accuracy_coic}
    \begin{scriptsize}
    \vspace{0pt}
    \begin{tabular}{|c|ccc|c|cc|c|}
    	\hline
        \multirow{2}{30pt}{\centering Method} &  \multicolumn{4}{c|}{Caltech-Office} & \multicolumn{3}{c|}{ImageNet-Caltech}\\
        \cline{2-8}
        & C 256 $\rightarrow$ W 10 & C 256 $\rightarrow$ A 10 & C 256 $\rightarrow$ D 10 & Avg & I 1000 $\rightarrow$ C 84 & C 256 $\rightarrow$ I 84 & Avg \\
        \hline
        AlexNet~\cite{cite:NIPS12CNN} & 58.44 & 76.64 & 65.86 & 66.98 & 52.37 & 47.35 & 49.86 \\
        DAN~\cite{cite:ICML15DAN} & 42.37 & 70.75 & 47.04 & 53.39  & 54.21 & 52.03 & 53.12\\
        RevGrad~\cite{cite:ICML15RevGrad} & 54.57 & 72.86 & 57.96 & 61.80 & 51.34 & 47.02 & 49.18\\
        RTN~\cite{cite:NIPS16RTN} & 71.02 & 81.32 & 62.35 & 71.56 & 63.69 & 50.45 & 57.07\\
        \hline
        SAN-selective & 76.44 & 81.63 & 80.25 & 79.44 & 66.78 & 51.25 & 59.02 \\
        SAN-entropy & 72.54 & 78.95 & 76.43 & 75.97 & 55.27 & 52.31 & 53.79\\ 
        SAN & \textbf{88.33} & \textbf{83.82} & \textbf{85.35} & \textbf{85.83} & \textbf{68.45} & \textbf{55.61} & \textbf{62.03}\\ 
    	\hline
    \end{tabular}
    \end{scriptsize}
    \vspace{-10pt}
\end{table}

The results reveal several interesting observations. \textbf{(1)} Previous deep transfer learning methods including those based on adversarial-network like RevGrad and those based on MMD like DAN perform worse than standard AlexNet, which demonstrates the influence of negative transfer effect. These methods try to transfer knowledge from all classes of source domain to target domain but there are classes in source domain that do not exist in the target domain, a.k.a. outlier source data. Fooling the adversarial network to match the distribution of outlier source data and target data will make the classifier more likely to classify target data in these outlier classes, which is prone to negative transfer. Thus these previous methods perform even worse than standard AlexNet. However, SAN outperforms them by large margins, indicating that SAN can effectively avoid negative transfer by eliminating the outlier source classes irrelevant to target domain. \textbf{(2)} RTN performs better than AlexNet because it executes entropy minimization criterion which can avoid the impact of outlier source data to some degree. But comparing RTN with SAN-selective which only has entropy minimization loss, we observe that SAN-selective outperforms RTN in most tasks, demonstrating that RTN also suffers from negative transfer effect and even the residual branch of RTN cannot learn the large discrepancy between source and target domain. SAN outperforms RTN in all the tasks, proving that our selective adversarial mechanism can jointly promote positive transfer from relevant source domain data to target domain and circumvent negative transfer from outlier source domain data to target domain.

We go deeper into different modules of SAN by comparing the results of SAN variants in Tables~\ref{table:accuracy_office} and ~\ref{table:accuracy_coic}. \textbf{(1)} SAN outperforms SAN-selective, proving that using selective adversarial mechanism can selectively transfer knowledge from source data to target data. It can successfully select the source data belonging to the classes shared with target classes by the corresponding domain discriminators. \textbf{(2)} SAN outperforms SAN-entropy especially in tasks where source and target domains have very large distribution gap in terms of the different numbers of classes, e.g. \textbf{I 1000} $\rightarrow$ \textbf{C 84}. Entropy minimization can effectively decrease the probability of predicting each point to irrelevant classes especially when there are a large number of irrelevant classes, which can in turn boost the performance of the selective adversarial mechanism. This explains the improvement from SAN-entropy to SAN.

\subsection{Analysis}

\begin{figure*}[!htbp]
  \centering
  \subfigure[Accuracy w.r.t \#Target Classes]{
    \includegraphics[width=0.32\textwidth]{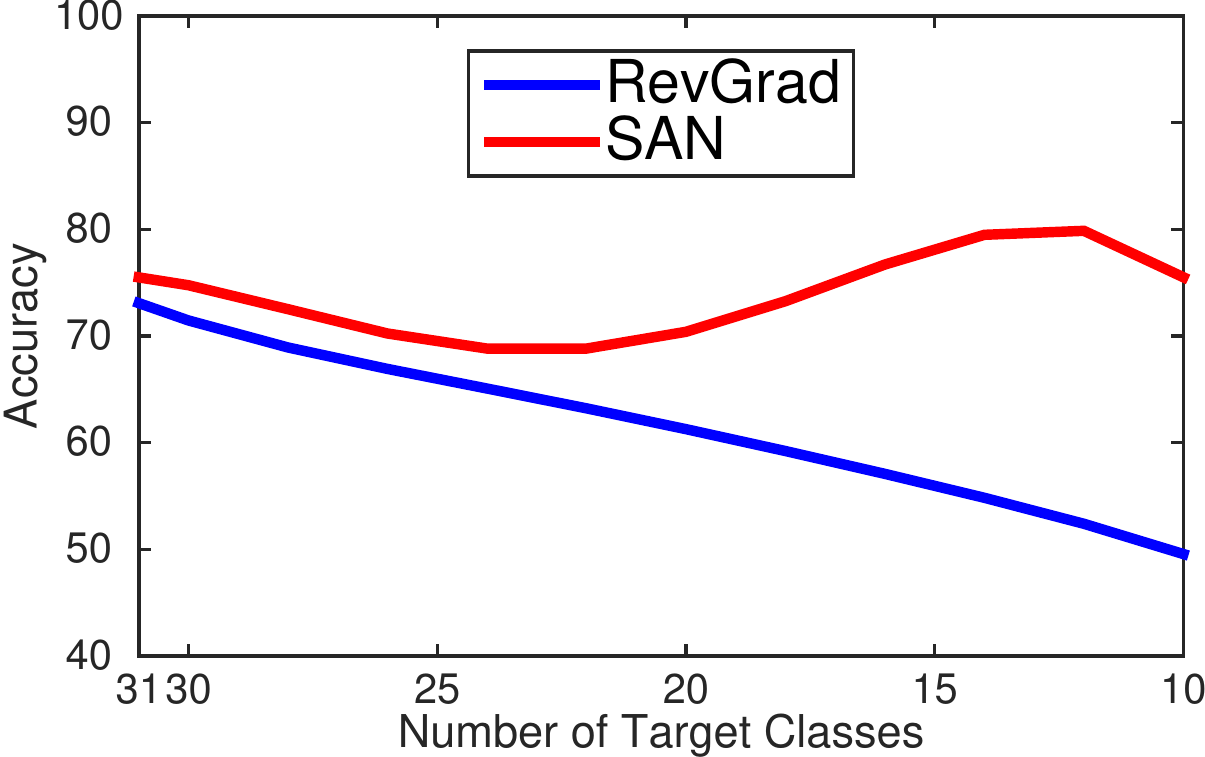}
    \label{fig:accuracy_number}
  }\hfil
  \subfigure[Test Error]{
    \includegraphics[width=0.33\textwidth]{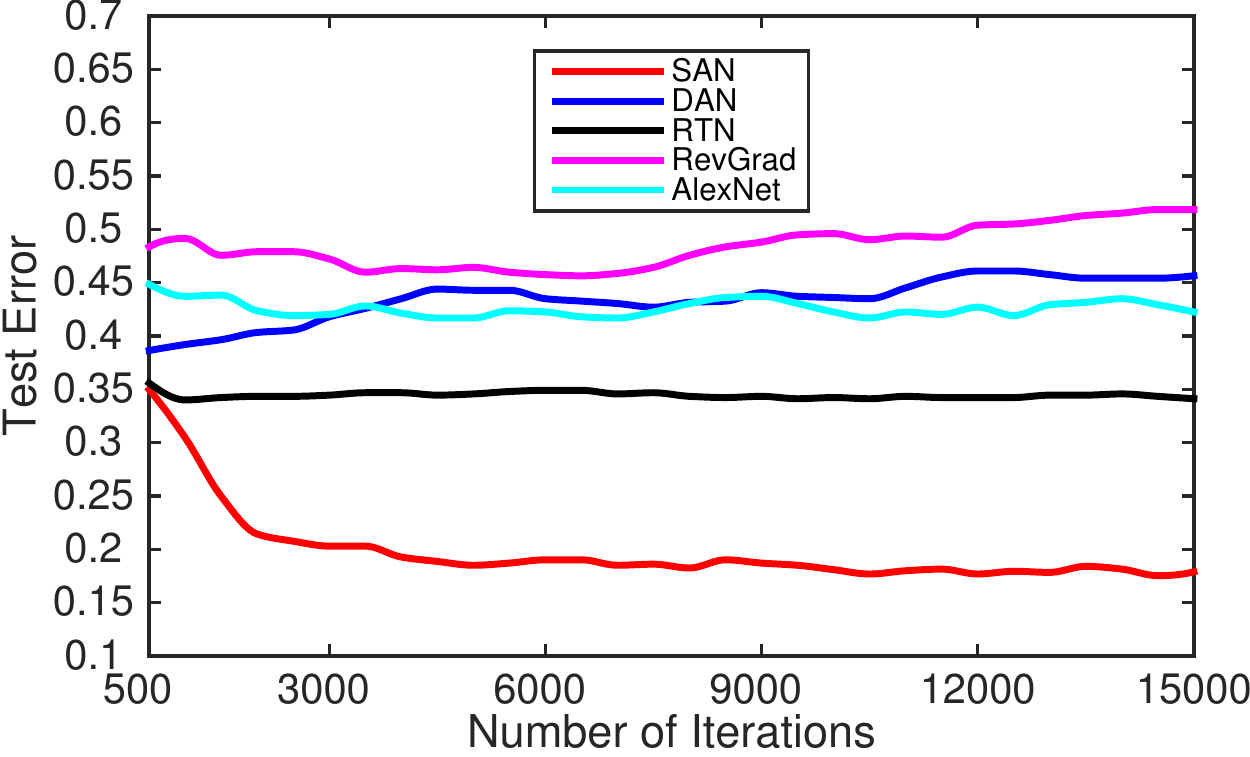}
    \label{fig:validation_error}
  }
  \vspace{-10pt}
  \caption{Empirical analysis: (a) Accuracy by varying \#target domain classes; (b) Target test error.}
  \vspace{-10pt}
\end{figure*}

\textbf{Accuracy for Different Numbers of Target Classes:}
We investigate a wider spectrum of partial transfer learning by varying the number of target classes. Figure~\ref{fig:accuracy_number} shows that when the number of target classes decreases, the performance of RevGrad degrades quickly, meaning that negative transfer becomes severer when the domain gap is enlarged. The performance of SAN degenerates when the number of target classes decreases from $31$ to $20$, where negative transfer problem arises but the transfer problem itself is still hard; the performance of SAN increases when the number of target classes decreases from $20$ to $10$, where the transfer problem itself becomes easier. The margin that SAN outperforms RevGrad becomes larger when the number of target classes decreases. SAN also outperforms RevGrad in standard transfer learning setting when the number of target classes is $31$.

\textbf{Convergence Performance:}
We examine the convergence of SAN by studying the test error through training process. As shown in Figure~\ref{fig:validation_error}, the test errors of DAN and RevGrad are increasing due to negative transfer. RTN converges very fast depending on the entropy minimization, but converges to a higher test error than SAN. SAN converges fast and stably to a lowest test error, meaning it can be trained efficiently and stably to enable positive transfer and alleviate negative transfer simultaneously. 

\begin{figure*}[!htbp]
  \centering
  \subfigure[DAN]{
    \includegraphics[width=0.19\textwidth]{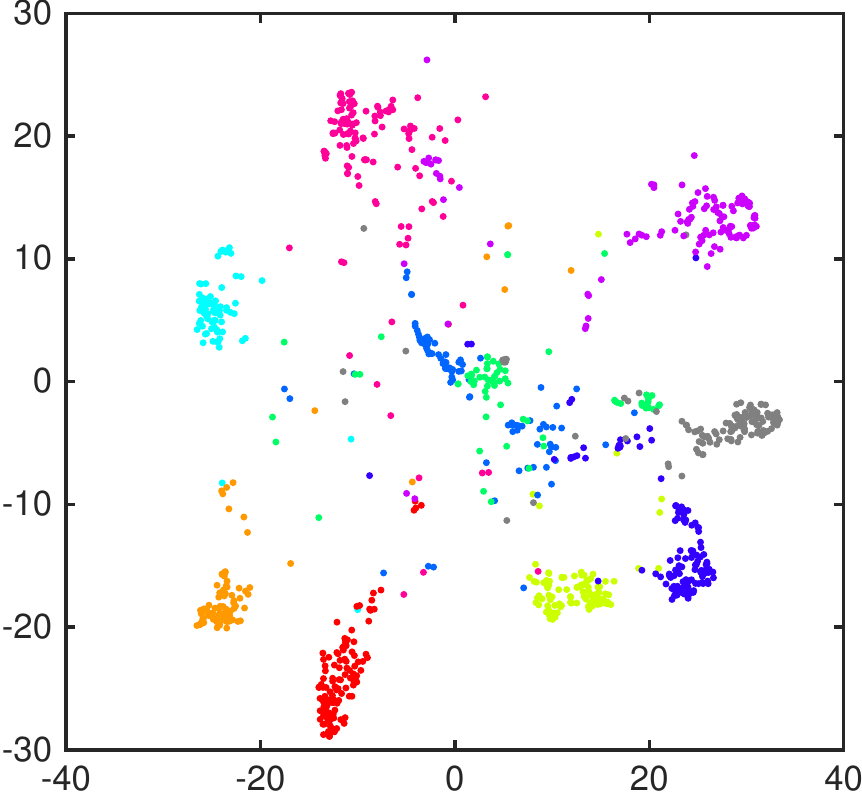}
    \label{fig:dan}
  }\hfil
  \subfigure[RevGrad]{
    \includegraphics[width=0.19\textwidth]{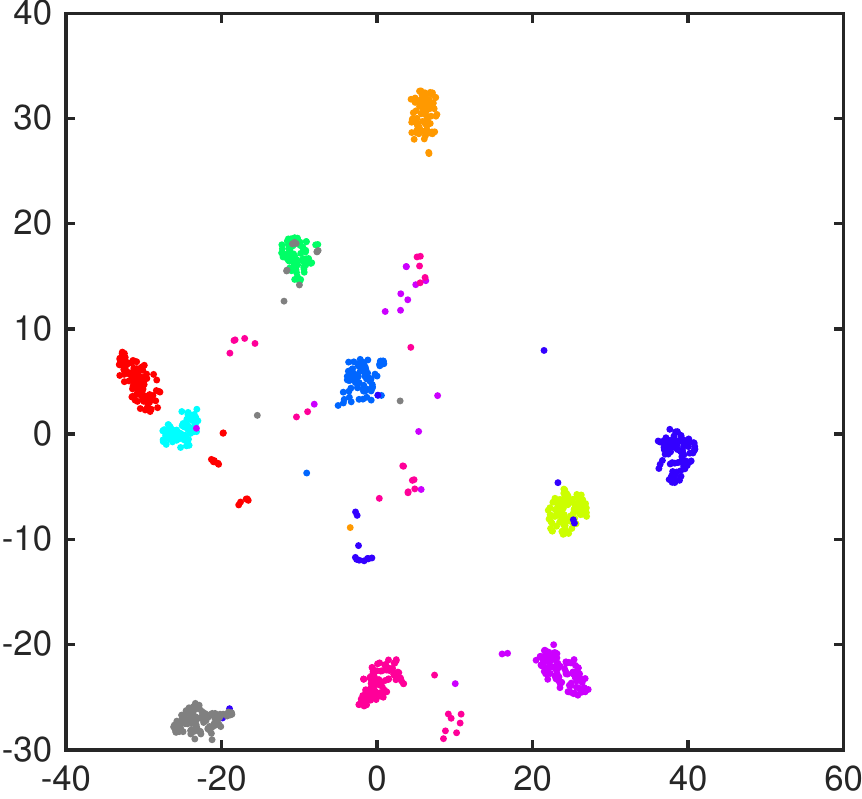}
    \label{fig:grl}
  }\hfil
  \subfigure[RTN]{
    \includegraphics[width=0.19\textwidth]{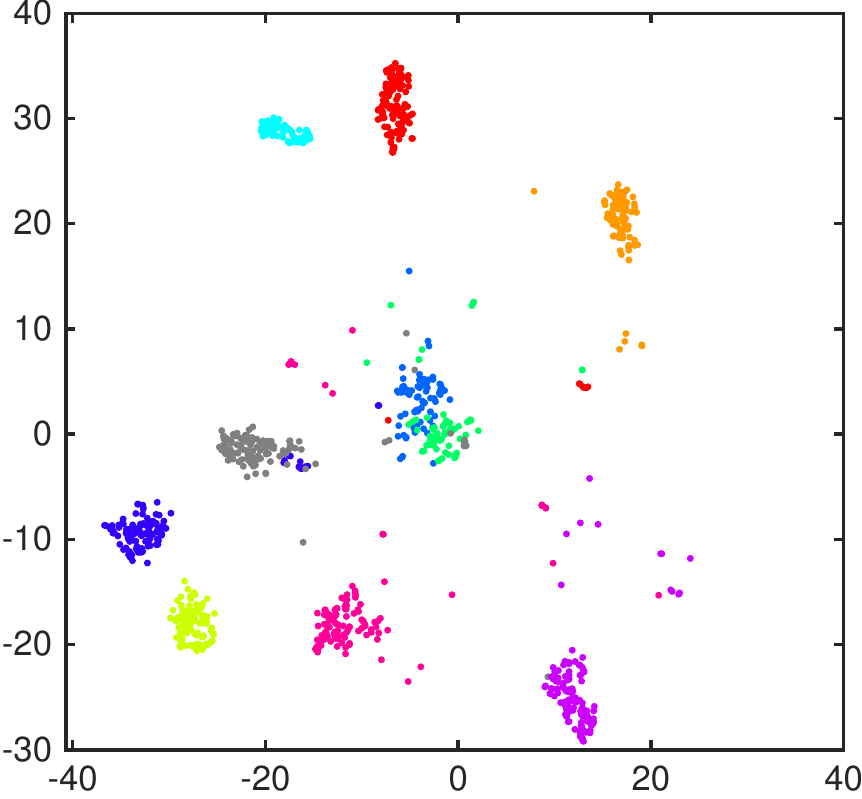}
    \label{fig:rtn}
  }\hfil
  \subfigure[SAN]{
    \includegraphics[width=0.26\textwidth]{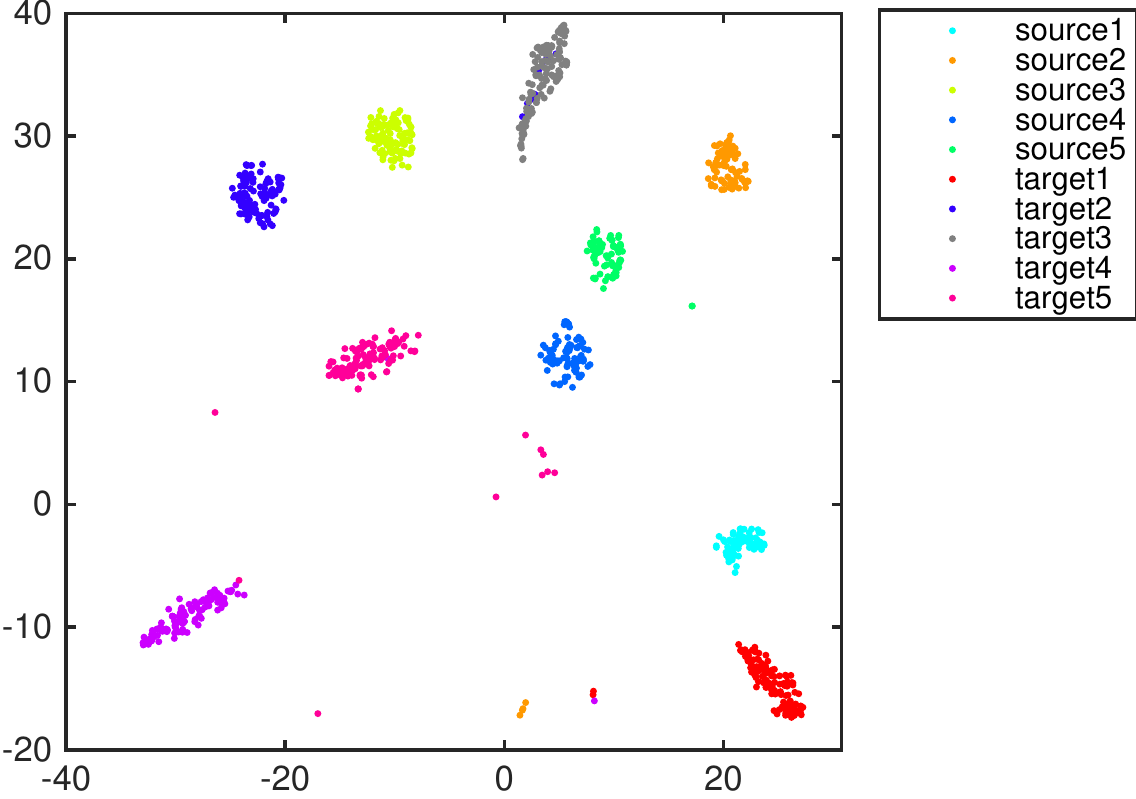}
    \label{fig:robust}
  }
  \vspace{-5pt}
  \caption{The t-SNE visualization of DAN, RevGrad, RTN, and SAN with class information.}
  \vspace{-10pt}
\end{figure*}

\begin{figure*}[!htbp]
  \centering
  \subfigure[DAN]{
    \includegraphics[width=0.19\textwidth]{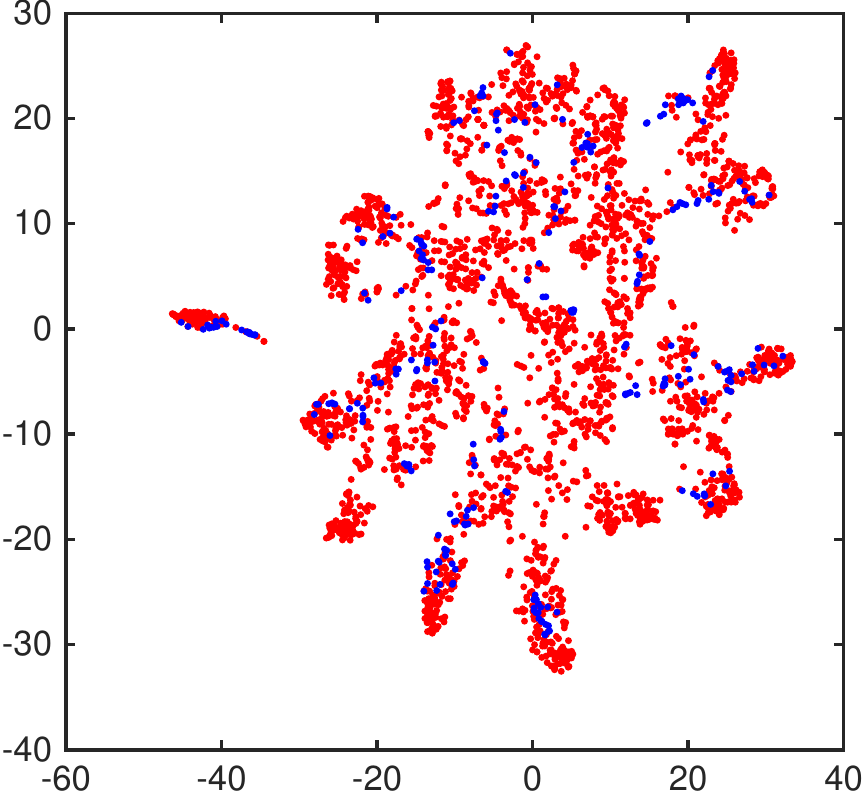}
    \label{fig:dan_st}
  }\hfil
  \subfigure[RevGrad]{
    \includegraphics[width=0.19\textwidth]{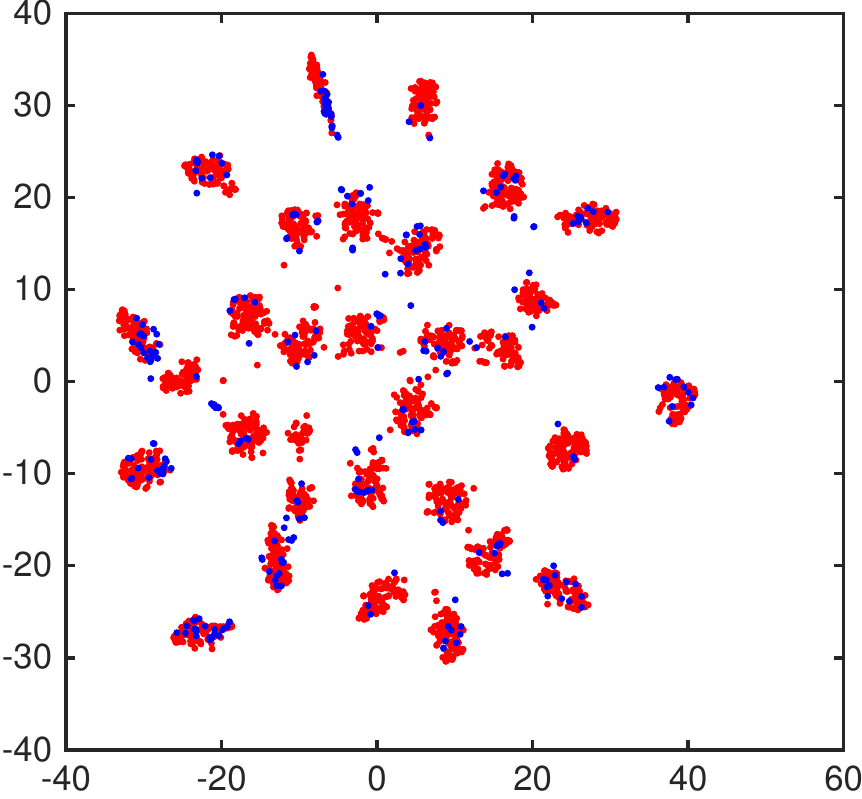}
    \label{fig:grl_st}
  }\hfil
  \subfigure[RTN]{
    \includegraphics[width=0.19\textwidth]{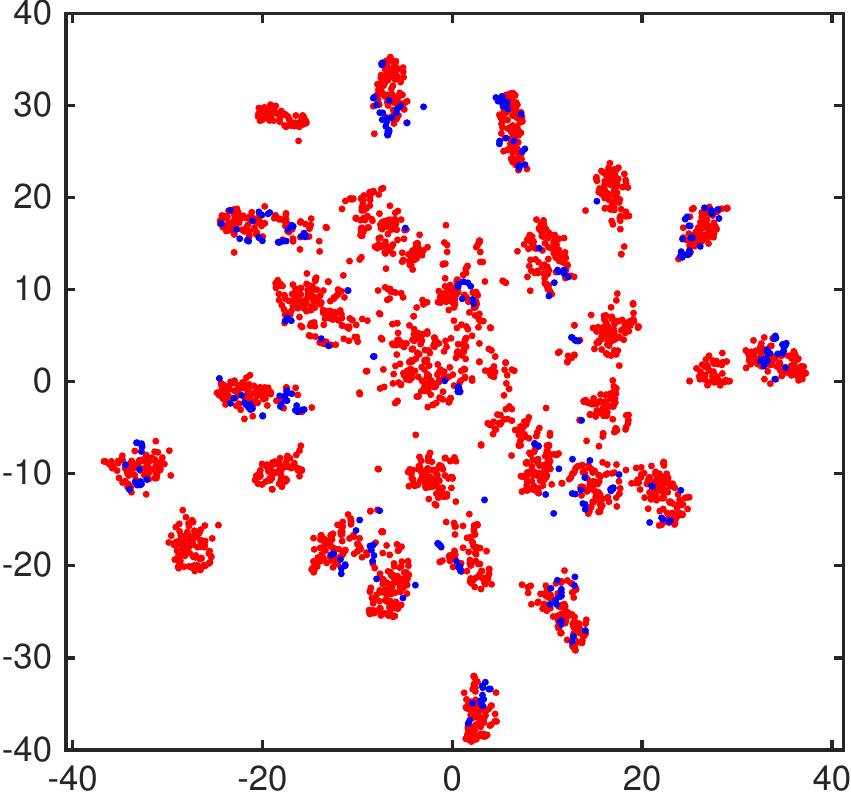}
    \label{fig:rtn_st}
  }\hfil
  \subfigure[SAN]{
    \includegraphics[width=0.26\textwidth]{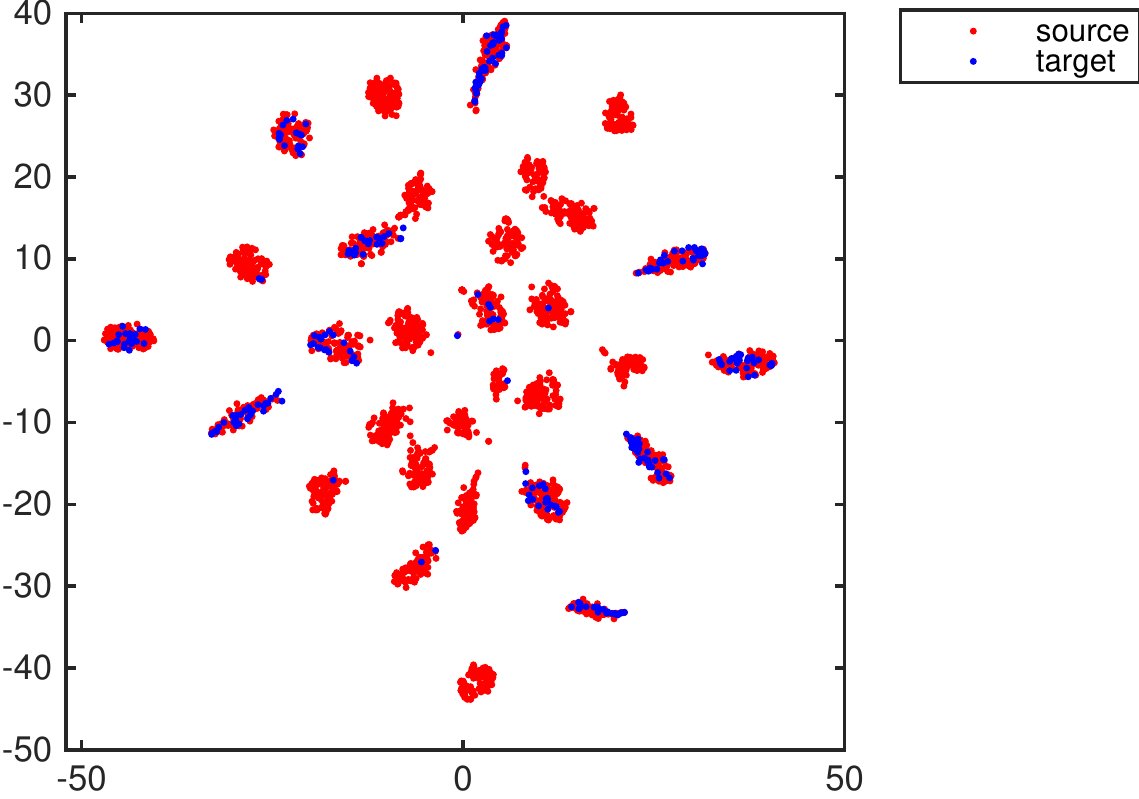}
    \label{fig:robust_st}
  }
  \vspace{-5pt}
  \caption{The t-SNE visualization of DAN, RevGrad, RTN, and SAN with domain information.}
  \vspace{-5pt}
\end{figure*}

\textbf{Feature Visualization:} We visualize the t-SNE embeddings~\cite{cite:ICML14DeCAF} of the bottleneck representations by DAN, RevGrad, RTN and SAN on transfer task \textbf{A 31} $\rightarrow$ \textbf{W 10} in Figures~\ref{fig:dan}--\ref{fig:robust} (with class information) and Figures~\ref{fig:dan_st}--\ref{fig:robust_st} (with domain information). We randomly select five classes in the source domain not shared with target domain and five classes shared with target domain. We can make intuitive observations. \textbf{(1)} Figure~\ref{fig:dan} shows that the bottleneck features are mixed together, meaning that DAN cannot discriminate both source and target data very well; Figure~\ref{fig:dan_st} shows that the target data are aligned to all source classes including those outlier ones, which embodies the negative transfer issue. \textbf{(2)} Figures~\ref{fig:grl}--~\ref{fig:rtn} show that both RevGrad and RTN discriminate the source domain well but the features of most target data are very close to source data even to the wrong source classes; Figures~\ref{fig:grl_st}--~\ref{fig:rtn_st} further indicate that both RevGrad and RTN tend to draw target data close to all source classes even to those not existing in target domain. Thus, their performance on target data degenerates due to negative transfer. 
\textbf{(3)} Figures~\ref{fig:robust} and \ref{fig:robust_st} demonstrate that SAN can discriminate different classes in both source and target while the target data are close to the right source classes, while the outlier source classes cannot influence the target classes. These promising results demonstrate the efficacy of both selective adversarial adaptation and entropy minimization. 

\section{Conclusion}
This paper presented a novel selective adversarial network approach to partial transfer learning. Unlike previous adversarial adaptation methods that match the whole source and target domains based on the shared label space assumption, the proposed approach simultaneously circumvents negative transfer by selecting out the outlier source classes and promotes positive transfer by maximally matching the data distributions in the shared label space. Our approach successfully tackles partial transfer learning where source label space subsumes target label space, which is testified by extensive experiments.

{\small
\bibliographystyle{ieee}
\bibliography{robust}
}

\end{document}